%% file: main.tex
\documentclass{article}

\usepackage{microtype}
\usepackage{graphicx}
\usepackage{subfigure}
\usepackage{booktabs} 
\usepackage{array,multirow,graphicx}

\usepackage{hyperref}

\usepackage[accepted]{icml2023}

\usepackage{amsmath}
\usepackage{amssymb}
\usepackage{mathtools}
\usepackage{amsthm}

\usepackage[makeroom]{cancel}
\usepackage{transparent}

\mathtoolsset{showonlyrefs}

\theoremstyle{plain}
\newtheorem{theorem}{Theorem}[section]

\theoremstyle{definition}

\theoremstyle{remark}
\newtheorem{remark}[theorem]{Remark}

\input{symbols.tex}

\usepackage[textsize=tiny]{todonotes}

\icmltitlerunning{Compute-lite Transformers}

\begin{document}

\twocolumn[
\icmltitle{LookupFFN: Making Transformers Compute-lite for CPU inference}



\icmlsetsymbol{equal}{*}

\begin{icmlauthorlist}
\icmlauthor{Zhanpeng Zeng}{wisc}
\icmlauthor{Michael Davies}{wisc}
\icmlauthor{Pranav Pulijala}{wisc}
\icmlauthor{Karthikeyan Sankaralingam}{wisc,nvidia}
\icmlauthor{Vikas Singh}{wisc}
\end{icmlauthorlist}

\icmlaffiliation{wisc}{University of Wisconsin, Madison, USA}
\icmlaffiliation{nvidia}{NVIDIA Research}

\icmlcorrespondingauthor{Zhanpeng Zeng}{zzeng38@wisc.edu}

\icmlkeywords{Machine Learning, ICML}

\vskip 0.3in
]



\printAffiliationsAndNotice{}  

\begin{abstract}
\input{sections/abstract.tex}
\end{abstract}

\section{Introduction}
\label{sec:introduction}
\input{sections/intro.tex}
\section{Preliminaries}
\input{sections/prelim.tex}

\section{FFN as Lookups}
\input{sections/formulation.tex}

\section{Experiments}
\input{sections/experiments.tex}

\section{Conclusions}
\input{sections/conclusions.tex}

\section*{Acknowledgments}
\input{sections/acknowledgement}

\bibliography{refs}
\bibliographystyle{icml2023}



\end{document}

%% file: symbols.tex
\newcommand{\R}{\mathbb{R}}

\newcommand{\norm}[1]{\left\|#1\right\|}

\newcommand{\T}{\top}

\DeclarePairedDelimiterX{\inner}[1]{\langle}{\rangle}{#1}

\newcommand{\Oh}{\mathcal{O}}

\newcommand{\entry}[1]{\left[#1\right]}

\newcounter{remark}

\newcommand{\vx}{\mathbf{x}}
\newcommand{\vy}{\mathbf{y}}
\newcommand{\vz}{\mathbf{z}}

\newcommand{\mA}{\mathbf{A}}
\newcommand{\mB}{\mathbf{B}}
\newcommand{\mC}{\mathbf{C}}
\newcommand{\mD}{\mathbf{D}}

\newcommand{\mF}{\mathbf{F}}

\newcommand{\mH}{\mathbf{H}}

\newcommand{\mR}{\mathbf{R}}
\newcommand{\mS}{\mathbf{S}}
\newcommand{\mT}{\mathbf{T}}

\newcommand{\mV}{\mathbf{V}}
\newcommand{\mW}{\mathbf{W}}
\newcommand{\mX}{\mathbf{X}}
\newcommand{\mY}{\mathbf{Y}}

%% file: sections/abstract.tex
While GPU clusters are the de facto choice for training large deep neural network (DNN) models today, several reasons including ease of workflow, security and cost have led to efforts investigating whether CPUs may be viable  for inference in routine use in many sectors of the industry. But the imbalance between the compute capabilities of GPUs and CPUs is huge. Motivated by these considerations, we study a module which is a workhorse within modern DNN architectures, GEMM based Feed Forward Networks (FFNs), and assess the extent to which it can be made compute- (or FLOP-) lite. Specifically, we propose an alternative formulation (we call it LookupFFN) to GEMM based FFNs inspired by the recent studies of using Locality Sensitive Hashing (LSH) to approximate FFNs. Our formulation recasts most essential operations as a memory look-up, leveraging the trade-off between the two resources on any platform: compute and memory (since CPUs offer it in abundance). For RoBERTa language model pretraining, our formulation achieves similar performance compared to GEMM based FFNs, while dramatically reducing the required FLOP. Our development is complemented with a detailed hardware profiling of strategies that will maximize efficiency -- not just on contemporary hardware but on products that will be offered in the near/medium term future. Code is avaiable at \url{https://github.com/mlpen/LookupFFN}.


%% file: sections/intro.tex
CPU-based inference in the data-center is growing in importance as evidenced by recent server chip announcements from IBM, Intel,  AMD and ARM~\cite{10.1145/3470496.3533042,intel-amx,9731107,zendnn,arm-linaro-cpu-server-ml} and academic efforts~\cite{10.5555/3358807.3358895,9410437,9499927,10.5555/3358807.3358897}. Some of the technical and business motivations include latency, security, privacy and the fact that modern data-intensive workloads have AI tasks embedded in a pipeline of non-AI tasks. 
Further, CPUs are a generic platform common across servers and clients faithfully serving 
the compute needs of businesses, which makes it attractive. 
And finally -- it comes down to the cost of running the 
full workload.
Unfortunately, CPU chips lack the computational intensity of raw-FLOPS compared to GPUs. 
On the positive side, CPUs provide tremendously large caches in the range of 128MB to 192MB, and even larger~\cite{9731678}, which is currently under-utilized. Furthermore, such caches made out of SRAMs are more than an order of magnitude more energy efficient to access compared to DRAMs (DDR, GDDR, or HBM)~\cite{9499913,6757323}, while providing $4\times$ access bandwidth increase~\footnote{AMD Zen2 for example, allows 64 bytes per cycle into each core: with 32 cores running 3.2 GHz that amounts to 6 TB/sec.}. In this context, this paper revisits one of the fundamental building blocks of modern deep-learning: the Feed Forward Network (FFN), to examine algorithmic reformulations to make it FLOP-lite and CPU friendly.

FFNs are essential components in almost all deep neural networks, such as convolutional networks \cite{howard2017mobilenets} or Transformers \cite{vaswani2017attention}. 
They heavily rely on General Matrix Multiply (GEMM), which is extremely compute intensive, especially for large scale models 
common in the community today. 
Many alternatives \cite{slide, mongoose, switch_transformer, shufflenet, acdc} have been proposed to reduce the FLOP needs of FFNs. 
In the context of 
inference, one may 
use generic pruning/quantization techniques \cite{NEURIPS2019_a01a0380, lee2018snip, Jacob_2018_CVPR, Q8BERT, I-BERT}, after 
training has concluded. This strategy is 
typically agnostic of specific modules in the architecture, 
and is applied to the entire model. Notice that if 
module-specific 
FLOP reduction is accomplished somehow, the full model will 
still benefit from a scheme like pruning, prior to deployment.  
Since such an idea would complement any 
reformulation of FFNs, it is more meaningful to instead 
focus the discussion on existing ideas that targets FLOP 
reduction further upstream. 

{\bf LSH can make FFNs FLOP-lite.}
One popular line of work shows how 
to use Locality Sensitive Hashing (LSH), 
to address the computational bottleneck of feed-forward via adaptive sparsity. 
For example, Slide \cite{slide} uses LSH to retrieve a small subset of units that omit high activation via maximum inner product search (MIPS) and only computes the outputs of these units, resulting in a sparse network. However, LSH poses 
certain difficulties. 
Due to the randomness of hash functions, a large number of hash functions are needed to get good MIPS results (see Fig. \ref{fig:lsh_issue}, left).
Also, the skewed hash bucket distribution makes LSH-based FFNs harder to parallelize since the computational load of different inputs can be very different (see Fig. \ref{fig:lsh_issue},  right). 
Considering training, due to the constantly evolving parameters, the hash tables need to be constantly updated to adapt for the changing parameter matrices (rehash) creating a large overhead during training \cite{mongoose}.

{\bf Extensions/modifications of Slide.} 
The Slide result shows that an 
approach based on LSH can be effective at 
reducing the FLOP count in FFNs. Motivated by this 
observation, the authors in Mongoose \cite{mongoose} 
proposed strategies for improvements. Specifically, by making 
the LSH component learnable and introducing a special update scheduler, 
Mongoose reduces the number of hash functions and the frequency of hash table updates (although the need for rehash is not eliminated).
Separately, the skewed bucket distribution still limits 
the parallelism of LSH. Nonetheless, 
the similarities with Slide in the use of LSH 
make Mongoose applicable for FLOP reductions in FFNs. 
A distinct use of LSH was shown in 
YOSO \cite{yoso} for approximating the self-attention 
matrix. Although a reduction in FLOP count did not motivate that 
work, the YOSO algorithm \cite{yoso} 
specifically adjusts LSH so that the skewness of bucket distribution does not affect the computational load -- to enable easy parallelism/efficiency gains. Therefore, 
YOSO is also potentially applicable for FLOP reductions in FFNs, like Slide/Mongoose.
However, a large number of hash functions (and to a lesser extent, rehash) cannot be avoided.
In summary, the effectiveness 
of these ideas for commonly used 
architectures remains unclear, 
which we will discuss in more detail later. 

{\bf This work and its contributions.} Motivated by the aforementioned limitations of LSH-based FFNs, in this paper, we provide a formulation of end-to-end learnable memory lookup for FFNs: LookupFFN. 
Specifically, we propose to directly view the hash tables as learnable modules. 
Projections are handled 
via a specialized module based on the fast 
Hadamard transform, which may be of independent interest. 
We show that the skewness of bucket distribution becomes irrelevant in our proposal. Since there are no parameter matrices as in \cite{mongoose, slide}, rehashing can be completely avoided, and the gradient updates are performed directly on the hash functions and hash tables 
(during model training).
The proposed formulation is differentiable, and no special optimization on the hash modules is needed. In practice, LookupFFN can simply be integrated into common DNN models, optimization flows, and software frameworks.


{\bf Main features.} 
Based on measurements and analytical calculations, we estimate 
{$6 \times$ (or more) reduction in FLOP} compared to a vanilla FFN with almost the same accuracy. 
Even though our formulation requires somewhat large tables (16MB and more), with careful algorithm design, we can make the access pattern somewhat cache-friendly -- {achieving nearly 80\% L1 cache hit rate}. In particular, hardware-managed caches work well, avoiding the need for excessive optimization for the software-managed shared-memory of a GPU. In practice, this means that we are able to reduce the {energy consumption of 80\% of access to SRAMs} (14 pico joules or pj per 64-bit access), versus 300 to 450 pj for DRAM-based access. We show that, on contemporary hardware, for inference, {we are 2.51$\times$ faster than a vanilla FFN}. With new technology like 3D caches appearing in CPUs, we expect SRAM based bandwidth to increase even further, making LookupFFN integer factors faster as memory technology and packaging improvements continue.
While this formulation has more memory lookups compared to GEMM-based FFN, the majority of the memory lookups are independent of each other. 
This means LookupFFN heavily relies on high memory throughput but has high tolerance to memory latency -- making designing software (and potentially hardware) implementation easier. 

%% file: sections/prelim.tex
First, we briefly review the Transformer and then the Feed-Forward Network (FFN) within the Transformer, the use case we study in this paper. Then, we will discuss related works that motivated this paper and some salient limitations. We use $\entry{\cdot}_i$ to denote the $i$-th row/entry of the matrix/vector, and use \textbf{BOLD} uppercase letters to denote matrices, \textbf{bold} lower case letters to denote vectors. 

\subsection{The Feed-forward Network in Transformer}

Given an embedding matrix $\mX \in \R^{n \times d}$ representing the embedding vectors of $n$ tokens, a Transformer layer is
\begin{align}
\mA &= \mathcal{G}(\mX) + \mX \\
\mY &= \mathcal{F}(\mA) + \mA
\end{align}
where $\mathcal{G}(\cdot)$ is a multi-head attention and $\mathcal{F}(\cdot)$ is a two layer FFN. There are layer normalizations within the Transformer layer, but for notational simplicity, we omit them. 
The focus of this paper is the efficiency (FLOP) of $\mathcal{F}(\cdot)$. 

\textbf{What about efficiency of $\mathcal{G}(\cdot)$?} There are various results studying how to improve the efficiency of $\mathcal{G}(\cdot)$ \cite{choromanski2020rethinking, xiong2021nystromformer, beltagy2020longformer, zaheer2020big, Kitaev2020ReformerTE, yoso, zeng2022mra}. 
Of course, conceptually, our method (discussed later) can be extended to multi-head attention. But this work would be a little redundant. Such functionality is available within YOSO \cite{yoso}, which provides a similar mechanism to support efficient self-attention calculation that works well but is inapplicable to FFN. Therefore, we focus on FFN $\mathcal{F}(\cdot)$. 

\textbf{FFN.} The $\mathcal{F}(\cdot)$ is a point-wise operation applied to each row of $\mA \in \R^{n \times d}$. Let $\vx \in \R^{d}$ be any row of $\mA$, $t$ be the hidden dimension, and $\mW$ and $\mV$ be two parameter matrices in $\mathcal{F}(\cdot)$. Then, a FFN is 
\begin{equation}
\vy := \mathcal{F}(\vx) = \sigma(\vx \mW^\T) \mV = \sum_{i=1}^{t} \sigma(\inner{\vx, \entry{\mW}_i}) \entry{\mV}_i
\label{eq:ffn}
\end{equation}
Here, biases are omitted. 
This operation is usually implemented via GEMM/matrix multiply. 
It takes $\Oh(d t)$ compute cost for one input and is usually the main bottleneck (in terms of FLOP) of a DNN. 





The authors in Slide \cite{slide} observed that when $\sigma$ is a softmax, the output of a FFN is dominated by only a few entries of $\sigma(\inner{\vx, \entry{\mW}_i})$ and proposed a sparse FFN, which uses LSH to perform a maximal inner product search (MIPS) among $\entry{\mW}_i$ for large $\sigma(\inner{\vx, \entry{\mW}_i})$ terms. Only activations of the search results $\mathcal{S}(\vx)$ are computed to approximate the full softmax with a reduced compute burden, 
\begin{equation}
\vy \approx \sum_{i \in \mathcal{S}(\vx)} \sigma(\inner{\vx, \entry{\mW}_i}) \entry{\mV}_i
\label{eq:lsh_approx}
\end{equation}
Constructing $\mathcal{S}(\vx)$ requires a pre-processing step that hashes $\entry{\mW}_i$ into multiple hash tables, and a querying step that hashes $\vx$ to these hash tables and collects all $\entry{\mW}_i$ from the buckets that $\vx$ is hashed to. There are some problems with this construction. 
\textbf{Rehashing: } Since $\entry{\mW}_i$ are constantly updated while training, the hash tables need to be constantly updated or re-constructed, referred to as rehashing. 
\textbf{Large \#-hashes: } The LSH relies on the randomness of hash functions, so a large number of hash functions are used to obtain accurate MIPS result resulting in high query time (see left plot of Fig. \ref{fig:lsh_issue}). 
\textbf{Bucket skewness:} The LSH bucket distribution is skewed, so the number of $\entry{\mW}_i$ in different buckets are quite different and there is no control of how many $\entry{\mW}_i$ will be hashed into one bucket (see right plot of Fig. \ref{fig:lsh_issue}). Therefore, $|\mathcal{S}(\vx)|$ varies for different inputs. This skewness makes the workload difficult to be parallelized. 

\input{figures/lsh_issues.tex}

Mongoose \cite{mongoose} proposed a scheduler to reduce the frequency of rehashing and learnable hash functions to learn data-dependent hashing. So, the number of hash functions can be reduced without sacrificing MIPS quality, thereby \cite{mongoose} partially reduces the \textbf{Rehashing} and \textbf{large \#-hashes} issues but introduces an additional auxiliary learning component for learnable hashing. Further, the \textbf{bucket skewness} still persists. 
\citet{yoso} proposed a method for approximating self-attention in Transformer models, which can be extended to approximating FFNs. 
\citet{yoso} shows that when $\sigma$ is similar to the collision probability of LSH, instead of keeping track of $S$ (as in Slide), one can store the summation of $\entry{\mV}_i$'s in hash buckets where $\entry{\mW}_i$ are hashed to, which is analogous to the LSH pre-processing step. Let $f_k$ be a hash function, and $\mT_k \in \R^{2^\tau \times d}$ be a hash table representing $2^\tau$ $d$-dimensional buckets.
\begin{equation}
\begin{split}
\entry{\mT_k}_j = \frac{1}{h} \sum_{f_k(\entry{\mW}_i) = j} \entry{\mV}_i \quad \vy \approx \sum_{k=1}^h \entry{\mT_k}_{f_k(\vx)}
\end{split}
\label{eq:yoso}
\end{equation}
Then, in the LSH querying step, we can directly estimate $\vy$ by computing an average of one bucket of multiple $\mT_k$ with a consistent compute cost. Therefore, the \textbf{bucket skewness} issue is solved. However, the steps of \cite{yoso} relies on the randomness of $f_k$, so it requires a large number of hash functions for a good estimate. Further, since $\mW$ and $\mV$ are evolving during training, $\mT_k$ needs to be recomputed after every parameter update, which is inefficient. The \textbf{rehashing} and \textbf{large \#-hashes} problems remain open. 

None of the foregoing methods  can resolve all issues. In particular, no method solves  \textbf{rehashing} -- all of them require rehashing when the parameters are updated. 
One of our goals is to completely eliminate the need for rehashing, 
and remove the dependency of workload on the bucket size. 
Further, we also hope to
obtain a scheme that, if desired, can be trained end-to-end via back-propagation.

\input{figures/method_comparison_visualization.tex}

%% file: figures/lsh_issues.tex
\begin{figure}[!tb]
\centering
\vspace{-0.05in}
\includegraphics[width=3.2in]{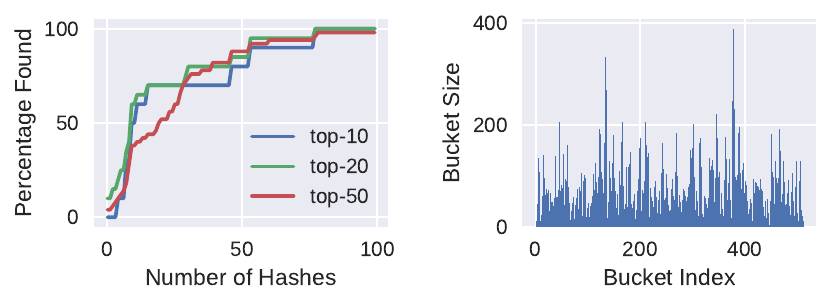}
\vspace{-0.15in}
\caption{The left plot shows the number of hash functions used versus the percentage of top-x nearest neighbors found using these hashes. A large number of hash functions are needed for accurate MIPS result. The query time is linearly proportional to the number of hash functions.
The right plot shows the bucket size of each bucket. It visualizes the bucket skewness issue. }
\label{fig:lsh_issue}
\vspace{-0.15in}
\end{figure}

%% file: figures/method_comparison_visualization.tex
\begin{figure*}[!t]
\centering
\vspace{-0.05in}
\includegraphics[width=6.5in]{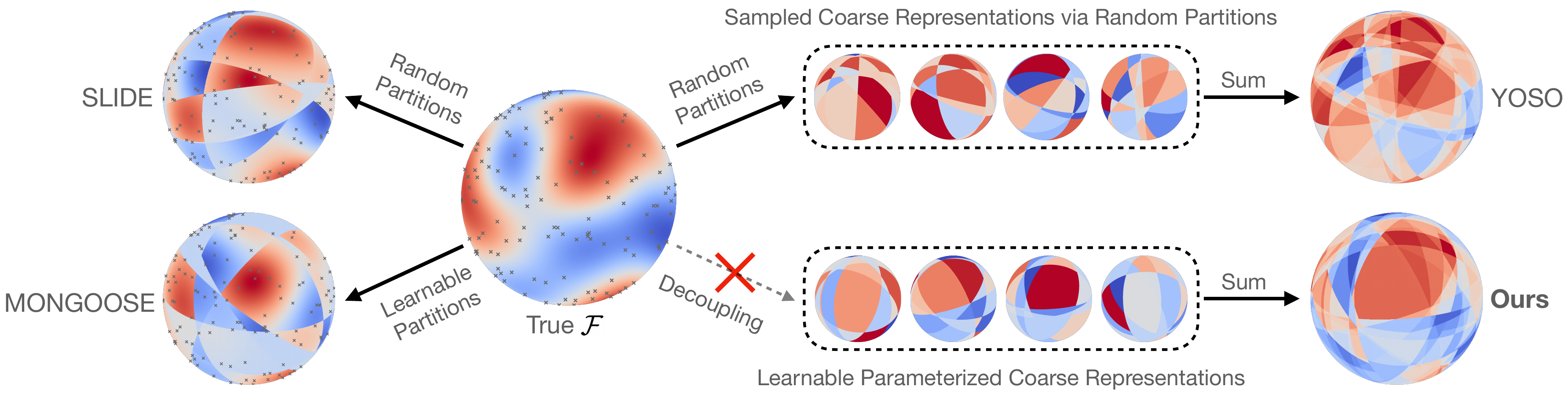}
\vspace{-0.05in}
\caption{High level comparison of each method. The true $\mathcal{F}(\cdot) = \sum_{i=1}^h \sigma(\inner{\cdot, \entry{\mW}_i}) \entry{\mV}_i$ is constructed as a function in $S^2$. Here, $\entry{\mW}_i \in S^2$ and $\entry{\mV}_i \in \R$. The points $\entry{\mW}_i$ are marked in the left three figures. 
SLIDE, MONGOOSE, and YOSO try to construct an approximation of the true $\mathcal{F}(\cdot)$ via different uses of LSH partitions, so whenever $\mathcal{F}(\cdot)$ is updated, the LSH partitions need to be updated. 
Rather than approximating the function $\mathcal{F}$, our proposed method is plugged into a deep learning model and uses the back-propagated gradient to learn appropriate transformation similar to a vanilla FFN. }
\label{fig:method_comparison_visualization}
\vspace{-0.2in}
\end{figure*}

%% file: sections/formulation.tex
Here, we present an end-to-end construction for differentiable table lookups as an efficient alternative to GEMM for FFNs where most operations are memory lookups. 

\subsection{Differentiable Lookup}
\label{subsec:lookup}

To avoid the impact of skewed bucket distribution on efficiency, we start from \eqref{eq:yoso} and attempt to adjust the formulation in the setting where it is used as a FFN. The randomness of $f_k$ in \eqref{eq:yoso} is the key ingredient of \cite{yoso}, but at the same time, this randomness introduces the need for a large number of hash functions to get an accurate approximation. This issue must be handled. Separately, we try to completely avoid any pre-processing steps or rehashing for evolving parameters $\mW$ and $\mV$. 

{\bf Main idea.} Observe that for YOSO, the $f_k$ is a partition of the $\R^d$ space and each hash table $\mT_k$ is a coarse representation of $\sum_{i=1}^h \sigma(\inner{\cdot, \entry{\mW}_i}) \entry{\mV}_i$ associated with $f_k$. Whenever $f_k$, $\mW$, or $\mV$ are updated, $\mT_k$ needs to be updated. This 
is inefficient. But $\mT_k$ is a coarse representation of a parameterized function, so we hypothesize that we might be able to directly optimize the coarse representation $\mH_k$ and $f_k$ to minimize the loss of the model. 
If possible, we also want to make it differentiable. If this is achieved, this strategy helps avoid any rehashing necessary in \eqref{eq:lsh_approx} and \eqref{eq:yoso}. Therefore, we consider the formulation
\begin{equation}
\begin{split}
{\transparent{0.4}\xcancel{\entry{\mT_k}_j := \frac{1}{h} \sum_{f_k(\entry{\mW}_i) = j} \entry{\mV}_i}} \quad \vy &:= \sum_{k=1}^h \entry{\mT_k}_{f_k(\vx)}
\end{split}
\label{eq:lookup}
\end{equation}
where $\mT_k$ and $f_k$ are learnable modules. Here, the dependency of $\mT_k$ on $f_k, \mW, \mV$, as in \eqref{eq:yoso}, is removed. 
Fig. \ref{fig:method_comparison_visualization} is a visualization of the difference comparing LSH-based FFNs. 
This decoupled dependency creates a problem in that the resultant formulation is not differentiable. \citet{yoso} uses the fact that \eqref{eq:yoso} is an estimate of a differentiable function, and uses the gradient of this function as an estimate of the gradient of \eqref{eq:yoso}, however, this estimate relies on the randomness of $f_k$ which is not available after decoupling in \eqref{eq:lookup}. So, the challenge is how we can train $f_k$ and $\mT_k$, and backpropagate to shallower layers. 

{\bf Making \eqref{eq:lookup} differentiable again.} To figure out a solution, we need to first dive into how $f_k$ is computed. \citet{yoso} uses the hyperplane hash \citep{charikar2002similarity} to compute the hash code. Specifically, define $\vz_k = \vx \mR_k$, referred to as the ``soft hash code'', where $\mR_k \in \R^{d \times \tau}$ is a random projection associated with $f_k$ where $\tau$ is the length of binary representation of the hash code. 
\begin{equation}
f_k(\vx) = \text{decimal}(\text{sign}(\vz_k))
\end{equation}
Here, $\text{decimal}$ is a function that maps the binary representation $\{\pm 1\}^{\tau}$ to a decimal representation $\{0, \cdots, 2^{\tau} - 1\}$. This form does not directly suggest a method for back-propagation, but observe that $f_k$ can be expressed as 
\begin{equation}
f_k(\vx) = \arg\max_i (\inner{\vz_k, \entry{\mS}_i})
\end{equation}
where $\mS \in \{\pm 1\}^{2^{\tau} \times \tau}$ is a structured matrix whose row vector 
\begin{equation}
\entry{\mS}_i = \text{decimal}^{-1}(i)
\end{equation}
is the binary representation of the integer $i$. While, by itself, this does not solve our problem, a common differentiable relaxation of $\arg\max$ is the softmax activation, and the resultant formulation for \eqref{eq:lookup} is, 
\begin{equation}
\hat{y}^* := \sum_{k=1}^h \sum_{i=1}^{2^\tau} \frac{\exp(\inner{\vz_k, \entry{\mS}_i}) \entry{\mT_k}_i}{\sum_{j=1}^{2^\tau} \exp(\inner{\vz_k, \entry{\mS}_j})}
\label{eq:full_softmax}
\end{equation}
Then, by replacing the random matrix $\mR_k$ with a learnable parameter matrix, this formulation makes $f_k$ a learnable hash function and $\mT_k$ a learnable coarse representation of a function in $\R^d$ in an end-to-end  manner. 

{\bf Remaining difficulties and solutions.} A naive implementation of this operation is extremely inefficient and has a runtime complexity of $\Oh(h 2^\tau d)$, which is not practical. A common choice of efficient softmax approximations is to use a small subset of softmax numerators (we denote the set of corresponding indices as $\mathcal{N}(\vz_k)$) to approximate the full softmax since the softmax is usually dominated by only a few entries within it \cite{spring2017new, charikar2017hashing}. Non-uniform sampling, such as LSH-based importance sampling, can be used to lower the estimation variance. However, we found that the structured matrix $\mS$ used in \eqref{eq:full_softmax} offers several properties that actually enables efficient approximation. Due to the structure of $\mS$, the denominator of \eqref{eq:full_softmax} can be rewritten as
\begin{equation}
\sum_{j=1}^{2^\tau} \exp(\inner{\vz_k, \entry{\mS}_j}) = 
 \prod_{j=1}^{\tau} (\exp(\entry{\vz_k}_j) + \exp(-\entry{\vz_k}_j))
\end{equation}
which only involves a $\Oh(\tau)$ cost. For calculating the numerator, we use a simple non-uniform sampling scheme for a better approximation of the softmax with a small number of samples. 
Due to the structure of $\mS$, we easily know the approximate sorting order of $\inner{\vz_k, \entry{\mS}_i}$ among different $i$. Specifically, note that
\begin{equation}
\begin{split}
\arg\max_{i} (\inner{\vz_k, \entry{\mS}_i}) &= \text{decimal}(\text{sign}(\vz_k)) \\
\arg\min_{i} (\inner{\vz_k, \entry{\mS}_i}) &= \text{decimal}(-\text{sign}(\vz_k))
\end{split}
\label{eq:min_max}
\end{equation}
When the order of magnitudes for different entries of $\vz_k$ are not too different, the $\norm{\entry{\mS}_i - \text{sign}(\vz_k)}_0$ term roughly indicates the magnitude of $\inner{\vz_k, \entry{\mS}_i}$. A smaller distance means a larger value.

\input{figures/diagram}

Therefore, we use the approximation
\begin{equation}
\hat{\vy} = \sum_{k=1}^h \sum_{i \in \mathcal{N}(\vz_k)} 
 \frac{ \exp(\inner{\vz_k, \entry{\mS}_i}) \entry{\mT_k}_i}{ \prod_{j=1}^{\tau} (\exp(\entry{\vz_k}_j) + \exp(-\entry{\vz_k}_j)) }
 \label{eq:partial_softmax}
\end{equation}
where $\mathcal{N}(\vz_k)$ can be easily sampled according to $\ell_0$ difference $\norm{\entry{\mS}_i - \text{sign}(\vz_k)}_0$. It is much easier compared to other non-uniform sampling based softmax approximations \cite{spring2017new, charikar2017hashing} since we can sample large numerators based on the number of sign flips away from $\text{sign}(\vz_k)$. Further, we empirically found that in most cases, just using the largest numerator, i.e., let
\begin{equation}
g(\vz_k) := \arg\max_{i} (\inner{\vz_k, \entry{\mS}_i})
\label{eq:optimal_index}
\end{equation}
computed via \eqref{eq:min_max}, then $\mathcal{N}(\vz_k) = \{g(\vz_k)\}$ is sufficient for  performance. This is our default choice for experiments. 

{\bf Two main operations.} The proposed learnable lookup consists of two operations: {\bf (a)} Hash: we compute multiple $\vz_k = \vx \mR_k$ for $k = 1, 2, \cdots, h$. Then, {\bf (b)} Gather: we use $g(\vz_k)$ (defined in \eqref{eq:optimal_index}) for memory lookup and calculate a weighted (based on $\vz_k$) accumulation of the lookup results $\entry{\mT_k}_{g(\vz_k)}$. This procedure is illustrated in Fig. \ref{fig:diagram}. 
 
\begin{remark}
While \eqref{eq:partial_softmax} might look unfamiliar, it is closely connected to two commonly used FFNs. When $\sigma$ in \eqref{eq:ffn} is the sigmoid activation, let $\vz_k = 0.5 \inner{\vx, \entry{\mW}_k}$, we note that \eqref{eq:ffn} can be rewritten as
\begin{equation}
y = \sum_{k=1}^h \frac{\exp(\vz_k) \entry{\mV}_k}{ \exp(\vz_k) + \exp(-\vz_k)} 
\end{equation}
which is just a special case of \eqref{eq:partial_softmax} with $\tau = 1$. When $\sigma$ is a GELU \cite{hendrycks2016gelu} commonly used in Transformer models, let $\vz_k = 0.851 \inner{\vx, \entry{\mW}_k}$, then a fast approximation of GELU can be written as
\begin{equation}
y = \sum_{k=1}^h \frac{1.175 \vz_k \exp(\vz_k) \entry{\mV}_k}{ \exp(\vz_k) + \exp(-\vz_k) } 
\end{equation}
This is again a special case of \eqref{eq:partial_softmax} with $\tau = 1$ and an additional linear scaling $1.175 \vz_k$. This scaling can be incorporated in \eqref{eq:partial_softmax} by an extra term $1.175 \inner{\vz_k, \entry{\mS}_i}$ in the numerator. 
\end{remark}


\subsection{BH4: Efficient and Expressive Projection}

\input{sections/projections.tex}

%% file: figures/diagram.tex
\begin{figure}[!tb]
\centering
\includegraphics[width=2.5in]{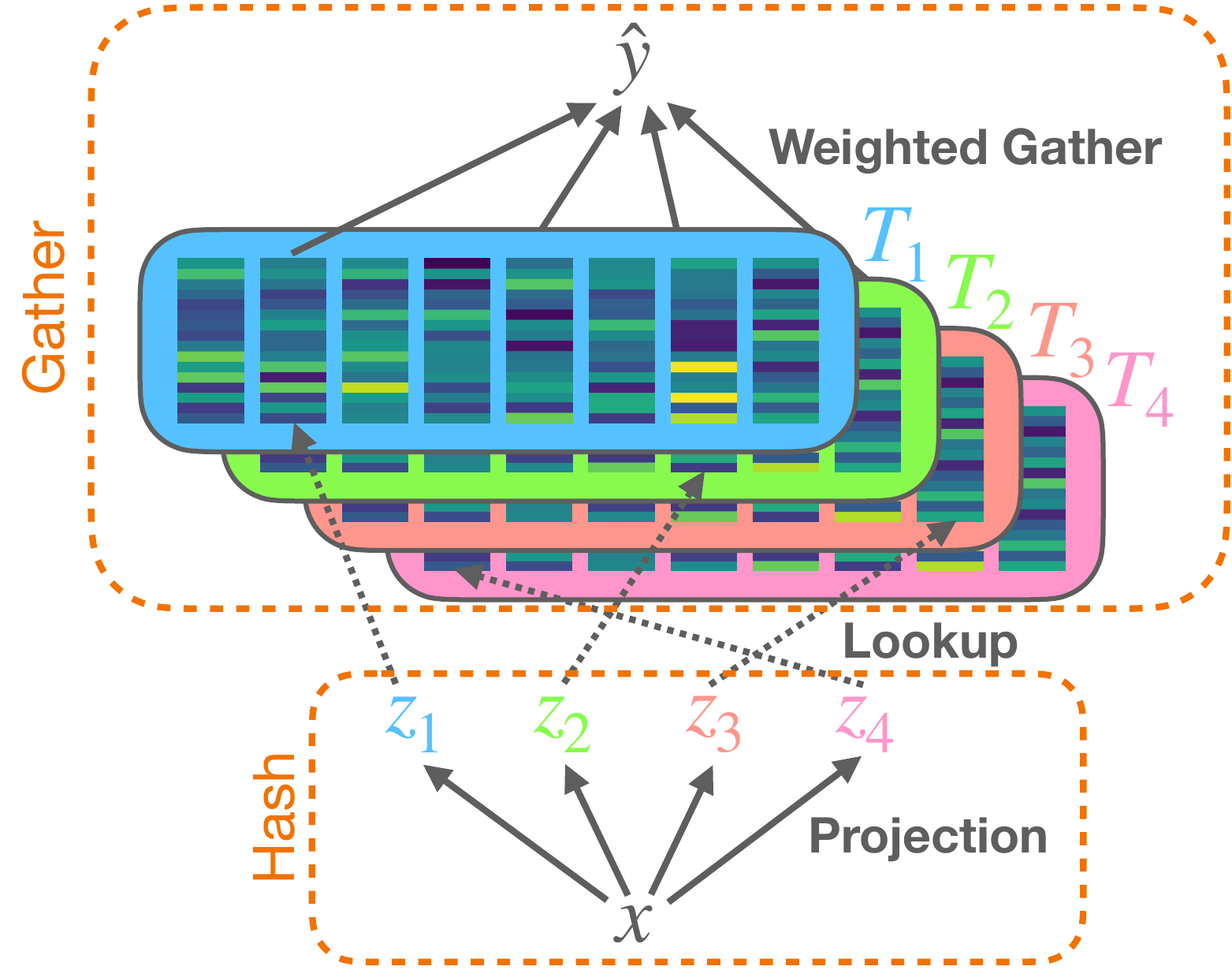}
\vspace{-0.15in}
\caption{Illustration of LookupFFN operations. }
\label{fig:diagram}
\vspace{-0.2in}
\end{figure}

%% file: sections/projections.tex
\label{subsec:bh4}

{\bf The problem.} In practice, we compute the ``soft hash code'' $\vz_k$ for multiple hash tables at once by computing $\vx \mR$
where $\mR \in \R^{d \times (h \tau)}$. Here, $\{\vz_1, \cdots, \vz_h\}$ are computed at once by partitioning $\vx \mR$ into $h$ $\tau$-dimensional vectors. 
The time complexity for this projection is $\Oh(h \tau d)$. This is not desirable since it is compute heavy. 

{\bf Some existing solutions yield unsatisfactory results.} For simplicity, we assume $h \tau = d$ and $d$ is power of $2$. When computing hash codes in the LSH setting, a common efficient alternative is efficient random projections implemented via a fast Hadamard transform with $O(d \log(d))$ cost. For example, \citet{yoso, alex2015practical} use
\begin{equation}
\mR := \mD_1 \mH \mD_2 \mH \mD_3 \mH
\label{eq:hadamard}
\end{equation}
where $\mD_i$ are matrices whose entries are $\{\pm 1\}$ for random sign flipping and $\mH$ is Hadamard transform. 
A simple learnable extension would be to replace $\mD_i$ with parameterized diagonal matrices. This belongs to a large family of structured efficient linear layers (SELLs) \cite{cheng2015exploration, fastfood, deepfried, acdc} For example, \citet{acdc} proposes a deep SELL, named ACDC, to increase the representation power:
\begin{equation}
\mR := \prod_{i=1}^k \mA_i \mC \mD_i \mC^{-1}
\label{eq:acdc}
\end{equation}
where $\mA_i, \mD_i$ are parameterized diagonal matrices and $\mC$ is the discrete cosine transform and $k$ is a hyper-parameter. A similar construction, but using the Hadamard transform would involve replacing $\mC$ with $\mH$. This can be viewed as a generalization of \eqref{eq:hadamard}. 
To empirically evaluate the representation power of each efficient projection, we use a toy problem of approximating a randomly generated matrix using these ideas. The results are shown in Fig. \ref{fig:projection_comparison}. 
We find that the representation power of \eqref{eq:acdc} and its Hadamard transform variant for small $k$ is extremely limited, but for large $k$, the efficiency is low and
the optimization becomes difficult. We can verify this optimization difficulty from the fact that as $k$ increases, the FLOP and parameter count increases, but the squared errors do not monotonically decrease.

\input{figures/projection_comparison.tex}

{\bf A simple yet highly effective scheme.} To address the optimization difficulty for large $k$, we propose that instead of scaling $k$, we can replace the diagonal matrices with their block diagonal counterparts, and scale the block size for the trade-off between efficiency and expressiveness. 
\begin{equation}
\mR := \prod_{i=1}^m \mB_i \mH
\label{eq:block_acdc}
\end{equation}
Here, $\mB_{i}$'s are parameterized block diagonal matrices with an adjustable block size. We refer to \eqref{eq:block_acdc} as BH\{m\} for different $m$. Fig. \ref{fig:efficient_projection_visualization} is a visualization of the projections discussed. 
We empirically verify that \eqref{eq:block_acdc} with $m = 4$ has a better trade-off between the expressiveness and efficiency compared to other $m$ values. When the FLOP or parameter counts are similar, larger $m$ does not increase its expressiveness, but a smaller $m$ decreases its expressiveness. 

\begin{remark}
Grouped convolution followed by channel shuffling in ShuffleNet \cite{shufflenet} can be directly expressed as $\mB \mF$: applying a block diagonal matrix followed by a structure transform $\mF$. ShuffleNet can be viewed as a BH1. We empirically verified that BH1 has near-identical approximation accuracy as ShuffleNet as shown in Fig. \ref{fig:projection_comparison}. 
\end{remark}

\input{figures/efficient_projection_visualization.tex}


%% file: figures/projection_comparison.tex
\begin{figure}[!tb]
\centering
\vspace{-0.1in}
\includegraphics[width=3.2in]{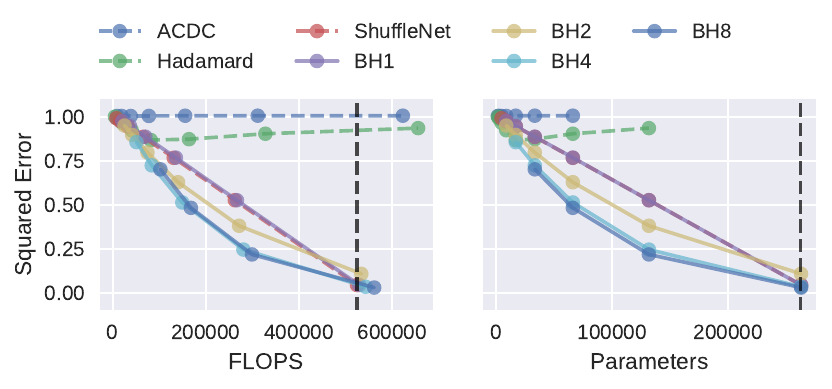}
\vspace{-0.15in}
\caption{Approximation capacity vs FLOPs and parameters for each efficient projections. Hadamard denotes a variant of ACDC by replacing discrete cosine transform with Hadamard transform. The vertical dash lines are the FLOPs and parameters of the vanilla projection. Any results to the right of the vertical dashed lines are not meaningful, as there is no efficiency gain. }
\label{fig:projection_comparison}
\vspace{-0.2in}
\end{figure}

%% file: figures/efficient_projection_visualization.tex
\begin{figure}[!t]
\centering
\includegraphics[width=3.2in]{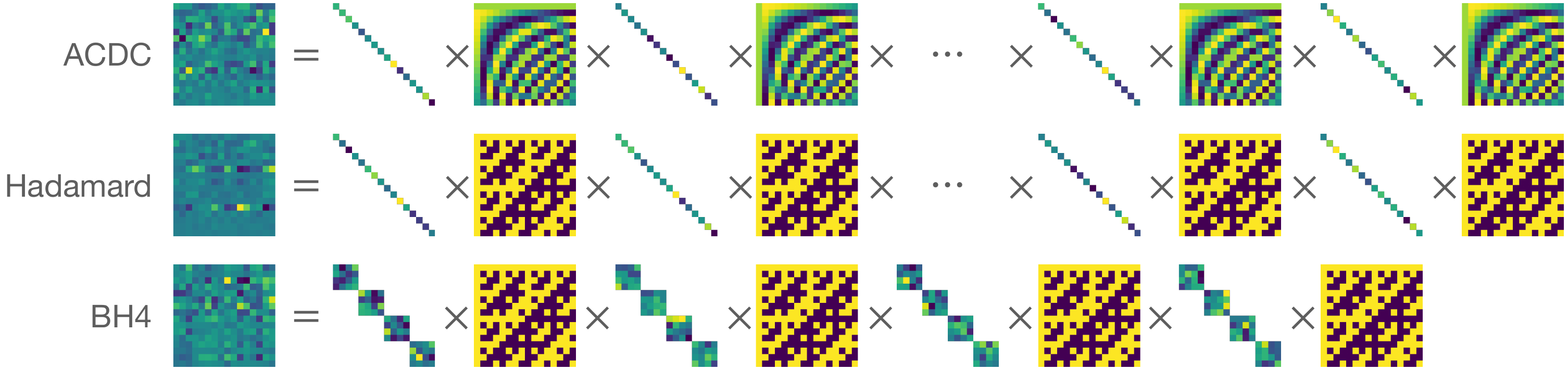}
\vspace{-0.1in}
\caption{Visualization of efficient projections. ACDC and its Hadamard variant increase their capacity by increasing the depth. BH4 increases its capacity by increasing its block size. }
\label{fig:efficient_projection_visualization}
\vspace{-0.17in}
\end{figure}

%% file: sections/experiments.tex
\label{sec:experiments}


In this section, we will present our empirical results evaluating the benefits/limitations of replacing VanillaFFN with a 
LookupFFN in a Transformer, and conduct 
a detailed performance profiling of LookupFFN. 

{\bf Note.} Slide and Mongoose report experiments on a two layer neural network whose last dimension is 200K (or even 670K) \cite{slide, mongoose}. This is 
a synthetic setting 
to extract peak practical speedup of FFN in Slide/Mongoose because 
more than 99\% of compute occurs in the final layer. 
In practical situations, such models are rare 
and might not reflect the actual workload of 
commonly used models, such as the Transformer. 
We are more interested in how these baselines and our method can be applied in more challenging commonly used DNN models, 
and what the corresponding performance impact is. 
As a result, we use the Transformer as a testbed to evaluate the effect of replacing VanillaFFN with baselines and our LookupFFN.

{\bf Outline.} In \S\ref{subsec:exp_performance}, we compare LookupFFN's performance and FLOP reduction to baselines, and check that our formulation scales without difficulty to full size (12-layer) models. 
Then, in \S\ref{subsec:exp_ablation}, to better understand its behavior, we perform an ablation study to study the effects of different hyper-parameters specific to lookup-based FFN. 
Finally, in \S\ref{subsec:exp_throughput}, we analyze the performance characteristics for LookupFFN. 
Since our goal is to reduce the FLOP count, for comparison or even individual assessment of LookupFFN, we include the estimated FLOP count next to the model performance for each table (for comparisons). FLOPs 
are estimated as the number of floating operations 
of processing a single instance (a single token in the context 
of a Transformer). 

Two variants are discussed in \S\ref{subsec:lookup} corresponding to two different activations. To align well with the GELU activation used in Transformer, we use the linearly scaled variant of \eqref{eq:partial_softmax} with $\mathcal{N}(\vz_k) = \{g(\vz_k)\}$: 
\begin{equation}
y = \sum_{k=1}^h 
 \frac{ \inner{\vz_k, \entry{\mS}_{g(\vz_k)}} \exp(\inner{\vz_k, \entry{\mS}_{g(\vz_k)}}) \entry{\mT}_{g(\vz_k)}}{ \prod_{j=1}^{\tau} (\exp(\entry{\vz_k}_j) + \exp(-\entry{\vz_k}_j)) }
\end{equation}

\textbf{Implementation Details.} We used PyTorch \citep{pytorch} for the majority of the implementation. On the GPU, our fast Hadamard Transform and weighted gather operators are not supported by PyTorch so we implemented custom CUDA kernels to support the operators for training. For CPU, we implemented these kernels in C++ using OpenMP for inference which uses AVX2 vector instructions.

\textbf{Evaluation Task.} For empirical evaluations, we use RoBERTa language modeling pretraining \cite{roberta} as our evaluation tool to measure the method performance, 
since it is a challenging task. 
The models are pretrained using masked language modeling \cite{devlin-etal-2019-bert} on the English Wikipedia corpus. We pretrain each model for 250K steps with a batch size of $256$, where each sequence is of length $512$. We use an Adam optimizer with 1e-4 learning rate, 10,000 warm-up steps, and linear decay. 
To keep compute reasonable, we use RoBERTa with 4 layers and 512 embedding dimensions for model evaluation except one stress-test experiment checking the scaling behavior of LookupFFN to a full size RoBERTa-base. 



\subsection{Performance Comparison}
\label{subsec:exp_performance}

\input{tables/baseline_comparison.tex}
\input{tables/base_models.tex}
\input{tables/large_table.tex}

\textbf{Comparing to Baselines.} We compare our method to Vanilla FFN, Slide \cite{slide}, Mongoose \cite{mongoose}, and YOSO \cite{yoso} based FFNs discussed previously. Additionally, for comparison to more baselines, we include Switch Transformer \cite{switch_transformer} and the grouped convolution + channel shuffling introduced in ShuffleNet. 
Others have identified that the original implementation of Slide, which is implemented from scratch in C++, is difficult to be adopted \cite{mongoose}, and have propose optimized variants, which we use here \cite{mongoose}. Instead of each instance in a batch retrieving its own subset of weights, the union of the retrieved subsets is used for computation. This strategy is used to avoid an irregular workload due to the skewed bucket distribution, as discussed earlier. The size of $\mathcal{S}(\cdot)$ is larger for larger batches. We note that in a Transformer model, the effective batch size for a FFN is the number of sequences $\times$ the sequence length. The union of retrieved subsets will simply contain the entire set of weights. 
For a more reasonable size of set $\mathcal{S}(\cdot)$, we partition the effective batch into smaller mini-batches (128 tokens for Slide and 2048 tokens for Mongoose) and feed them into the FFN sequentially. This would severely increase the runtime of training on GPUs. The size of mini batch is set such that it is small enough but running the experiment is still feasible. Further, since the performance of Slide \cite{slide}, Mongoose \cite{mongoose}, and YOSO \cite{yoso} based FFN largely depends on the frequency of rehashing, we perform rehashing after every parameter updates (this overhead is not counted towards FLOP) to ensure their optimal performance. 
The results are summarized in Tab. \ref{tab:baselines}. 
Our method achieves lower perplexity using fewer FLOP compared to the baselines. 
Further, the FLOP of our method can be significantly reduced with some loss in performance (but still better than the baselines except for the VanillaFFN) as shown in the last row of Tab. \ref{tab:baselines}. Additional results are discussed in \S\ref{subsec:exp_ablation}.

\input{tables/downstream}

\textbf{Downstream finetuning.} Further, we evaluate the quality of the pretrained language models for VanillaFFN and our LookupFFN on MNLI downstream task \citep{williams2018broad} in the GLUE benchmark \citep{wang2019glue}. The result is shown in Tab. \ref{tab:downstream}. We note that there is a small gap between our method and vanilla FFN, but the FLOP of our method is much lower. 

\textbf{Scaling to Full Size Models.} We check whether LookupFFN scales to a larger model, so we evaluate our method on a RoBERTa-base pretraining. All pretraining hyper-parameters remain the same as before. Due to the compute burden of training a full size model, we only perform one experiment comparing LookupFFN with $h = 170$ and $\tau = 9$ to VanillaFFN. The results are shown in Tab. \ref{tab:base_models}. Our method achieves $6.8 \times$ FLOP reduction while the log perplexity is only higher by $0.04$ in a RoBERTa-base model. 

\textbf{Memory use.} 
Our LookupFFN requires more memory (for large $h$ and $\tau$) since we directly parameterize the hash tables, but we believe this is not a key issue. In a GPU setting, memory use is critical since the GPU memory is usually much more expensive and limited. On the other hand, CPU memory is much cheaper and larger compared to GPU memory.

\subsection{Ablation Study}
\label{subsec:exp_ablation}

\textbf{Reducing FLOP for Hash.} 
We note that the projection in the Hash step has a complexity of $O(h \tau d)$ when using a vanilla dense projection and will generate the majority of FLOP for our LookupFFN. In \S\ref{subsec:bh4}, we propose an efficient alternative, BH4 and verify that the block size of $B_i$ has a direct impact on the representation power of BH4.
But will this impact the final performance of a model? To evaluate the trade-off between efficiency and performance, we study the effect of block size of $B_i$ on the log perplexity of RoBERTa. The results are summarized in the top table of Tab. \ref{tab:large_table}. 
When using a vanilla projection, the FLOP for the Hash step accounts for $89\%$ of the total FLOP. It is critical to reduce the FLOP need 
for this projection, else it becomes the main bottleneck. 
When using BH4, the performance decreases and efficiency increases as the block size decreases, which is expected, but it is surprising that while offering a large FLOP reduction, the performance drop compared to the vanilla dense projection is quite small. 

\textbf{Scaling Effect of LookupFFN.} 
The number of hash tables $h$ and the length of the hash code $\tau$ (or log of table size) control the scaling behavior. 
As a preliminary step, we first verify that our method can indeed be scaled for better accuracy by increasing the number of hash tables $h$. Therefore, we compare the model perplexity for different $h$ when $\tau$ is fixed, as shown in Tab. \ref{tab:large_table} (middle). The model performance monotonically increases as $h$ increases. When $h = 256$, our method achieves a lower perplexity compared to the VanillaFFN in Tab. \ref{tab:baselines}. 
Then, we evaluate the effect of scaling $\tau$. Instead of fixing $h$ while scaling $\tau$, we increase $\tau$ but at the same time reduce $h$ such that $h \tau$ is roughly the same. As shown in Tab. \ref{tab:large_table} (bottom), this comparison reveals a surprising scaling effect of our method: when $h\tau$ is fixed (the FLOP of Hash step remains the same), by increasing $\tau$ (increasing the table size), we can reduce the number of hash tables and reduce the 
FLOP count for the Gather step while achieving better performance.



\subsection{Throughput and Latency Study}

\label{subsec:exp_throughput}

In this subsection, we isolate the FFN from RoBERTa to examine the runtime and throughput of each style of FFN. We compare throughput of LookupFFN, Slide and Mongoose to VanillaFFN (YOSO is excluded since it does not have a CPU implementation at this time), then provide an in-depth analysis of LookupFFN's hardware-level behavior and potential for future throughput scalability. For all empirical results here, we use batch size 64 and sequence length 512, so the effective batch size is 32768 (side note: multi-head attention takes $274$ms in this setting).

\input{tables/runtime_compare.tex}

\textbf{Latency Comparison.} In order to demonstrate the latency improvement afforded by our LookupFFN for CPU inference, we compare runtime to alternatives. Tab. \ref{tab:throughput} shows the average per-iteration time for vanilla, Slide, and Mongoose-based FFN which is sized to match typical hyperparameters for a standard Transformer model on a modern AMD EPYC-7452 (Zen 2) 32-core Server. Our basic implementation for LookupFFN uses OpenMP without additional software-engineering optimization along with a naive implementation of BH4 and achieves 23\% speedup over VanillaFFN. We did further optimization of both the hash function and gather operation (\textit{opt1}), {achieving $2.51\times$ speedup} over VanillaFFN. Overall we see that both Slide and Mongoose perform worse than vanilla, while our optimized LookupFFN provides good performance improvement.

\textbf{Discrepancy between FLOP and Latency.} We note that there is difference in the speed up between FLOP and Latency in Tab. \ref{tab:baselines} and Tab. \ref{tab:throughput}. The latency not only depends on FLOP, but also the memory access pattern or arithmetic intensity. The GEMM used in VanillaFFN has a very structured memory access pattern and its arithmetic intensity is high, but the gather operator used in LookupFFN has a more random memory access pattern that depends on the input and its arithmetic intensity is lower. As a result, while the FLOP of LookupFFN is drastically lower than VanillaFFN, the corresponding speed up in latency is not as drastic.

\input{tables/runtime_analysis.tex}

\textbf{Performance analysis.} To further elucidate how we are able to achieve speedup, we detail architectural level performance statistics in Tab. \ref{tab:analysis} 
We notice the speedup primarily is afforded by achieving a higher sustained cache bandwidth. The lower LLC miss rate suggests this is at least in part due to extracting more reuse from the LLC, thereby making internal cache management including MSHRs, and on-chip network traffic perform better. 



\input{figures/speedup_opp.tex}

\textbf{Future Performance Opportunities.} Future performance gains for LookupFFN are possible through a combination of software and upcoming hardware optimizations which we summarize in Fig. \ref{fig:speedup_opp}. \textit{naive} and \textit{opt1} are our two configurations reported previously. Additional tuning of block size for the BH4 projection (64 to 16) could provide a $4\times$ speedup for our Hash step (\textit{hash-opt2}). Future cache technology such as 3D stacking \cite{9731565} will likely provide a rather generous boost in high-speed LLC cache capacity, enabling larger table size LookupFFN weights to be entirely cache-resident. In combination with this, careful cache optimization through the use of modern prefetching techniques \cite{10.5555/3291656.3291744}, as well as hardware improvements such as Intel's wide 128 B L1 cache interface \cite{9747991} can improve hit rates and overall bandwidth. If 90\% of bandwidth from a 128 B cache interface could be sustained, Gather step can be improved by a factor of 35$\times$ (\textit{gather-opt2}). Overall, these improvements could yield {$8.49\times$} improvement over VanillaFFN. Datatype precision reduction could potentially afford a further multiplicative runtime improvement of $2\times$, achieved by switching from float32 to float16 -- however, VanillaFFN would also gain a $2\times$ execution time reduction from float16.


\textbf{Discussion of Compiler Techniques for FFN.} Deep Learning specific compilers, such as TVM and XLA (built into TensorFlow) have been introduced, aiming to optimize operations such as a feed-forward network. We evaluated TVM and XLA on VanillaFFN to compare LookupFFN's latency to these two compiler frameworks. TVM performs 21\% worse than PyTorch, with a latency of $488$ms. Switching to TensorFlow gives $2.39\times$ improvement over the PyTorch VanillaFFN, with XLA yielding an additional 37\% performance boost (absolute latency of TF+XLA: $123$ms). Our optimized LookupFFN already provides {competitive performance} compared to TensorFlow, 
and {our additional hash optimization, when fully implemented/integrated, LookupFFN will provide performance improvement over TF+XLA.}




%% file: tables/baseline_comparison.tex
\begin{table}[!tb]
\begin{center}
\setlength{\tabcolsep}{4pt}
\begin{small}
\begin{tabular}{lcccc}
\toprule
Method & $h$ & $\tau$ & MFLOP & Log Perplexity \\
\midrule
VanillaFFN & - & - & 4.19 & 1.78 \\
Switch Transformer & - & - & 2.11 & 1.85 \\
Channel Shuffle & - & - & 2.10 & 1.96 \\
Slide & - & - & 1.32 & 1.98 \\
Mongoose & - & - & 3.21 & 1.87 \\
YOSO & - & - & 0.35 & 2.13 \\
\midrule
LookupFFN & 256 & 8 & 1.38 & 1.74 \\
       & 128 & 8 & 0.69 & 1.81 \\
\bottomrule
\end{tabular}
\end{small}
\end{center}
\vspace{-0.1in}
\caption{Log perplexity of each baseline. (lower is better) LookupFFN was tested with two different hyper-parameter configurations specified in $h$ and $\tau$ columns. }
\label{tab:baselines}
\vspace{-0.05in}
\end{table}

%% file: tables/base_models.tex
\begin{table}[!tb]
\begin{center}
\begin{small}
\begin{tabular}{lccccc}
\toprule
Method & $h$ & $\tau$ & MFLOP & Log Perplexity \\
\midrule
VanillaFFN & - & - & 9.44 & 1.37 \\
LookupFFN & 170 & 9 & 1.39 & 1.41 \\
\bottomrule
\end{tabular}
\end{small}
\end{center}
\vspace{-0.1in}
\caption{Log perplexity when scaling to a RoBERTa-base model.}
\label{tab:downstream}
\vspace{-0.05in}
\end{table}

%% file: tables/large_table.tex
\begin{table}[!th]
\begin{center}
\begin{small}
\setlength{\tabcolsep}{4pt}
\begin{tabular}{lccccc}
\toprule
\multicolumn{2}{l}{Type} & Block Size & \multicolumn{2}{c}{MFLOP} & Log Perplexity \\
& & & \scriptsize{Hash} & \scriptsize{Gather} & \\
\midrule
\multicolumn{2}{l}{Dense} & - & 1.05 & 0.13 & 1.79 \\
\multicolumn{2}{l}{BH4} & 64 & 0.56 & 0.13 & 1.81 \\
\multicolumn{2}{l}{BH4} & 32 & 0.30 & 0.13 & 1.83 \\
\multicolumn{2}{l}{BH4} & 16 & 0.17 & 0.13 & 1.85 \\
\bottomrule
\toprule
$h$ & $\tau$ & $h \tau$ & \multicolumn{2}{c}{MFLOP} & Log Perplexity \\
\midrule
32 & 8 & 256 & \multicolumn{2}{c}{0.31} & 1.94 \\
64 & 8 & 512 & \multicolumn{2}{c}{0.35} & 1.88 \\
128 & 8 & 1024 & \multicolumn{2}{c}{0.69} & 1.81 \\
256 & 8 & 2048 & \multicolumn{2}{c}{1.38} & 1.74 \\
\bottomrule
\toprule
$h$ & $\tau$ & $h \tau$ & \multicolumn{2}{c}{MFLOP} & Log Perplexity \\
& & & \scriptsize{Hash} & \scriptsize{Gather} & \\
\midrule
64 & 4 & 256 & 0.28 & 0.07 & 1.98 \\
32 & 8 & 256 & 0.28 & 0.03 & 1.94 \\
20 & 13 & 260 & 0.28 & 0.02 & 1.87 \\
\bottomrule
\end{tabular}
\end{small}
\end{center}
\vspace{-0.1in}
\caption{Ablation study evaluating the effects of different hyper-parameters on model performance. The MFLOP columns in the top and button tables are broken down into two column showing the FLOP for Hash and Gather steps separately. }
\label{tab:large_table}
\end{table}

%% file: tables/downstream.tex
\begin{table}[!tb]
\begin{center}
\begin{small}
\begin{tabular}{lccccc}
\toprule
Method & $h$ & $\tau$ & MFLOP & MNLI-m/mm \\
\midrule
VanillaFFN & - & - & 4.19 & 75.0/76.3 \\
LookupFFN & 256 & 4 & 0.82 & 74.1/74.7 \\
\bottomrule
\end{tabular}
\end{small}
\end{center}
\vspace{-0.1in}
\caption{Downstream performance of RoBERTa-small models. }
\label{tab:base_models}
\vspace{-0.15in}
\end{table}

%% file: tables/runtime_compare.tex
\begin{table}[!tb]
\begin{center}
\begin{small}
\begin{tabular}{lcc}
\toprule
Technique & Avg. Latency (ms) & Speedup  \\
\midrule
VanillaFFN    & 403 & $1.00\times$ \\
SLIDE      & 428 & $0.94\times$ \\
Mongoose   & 878 & $0.46\times$ \\
LookupFFN     & 328 & $1.23\times$ \\
LookupFFN (Opt1) & {160} & {$2.51\times$} \\
\bottomrule
\end{tabular}
\end{small}
\end{center}
\vspace{-0.1in}
\caption{Average latency for LookupFFN compared to baselines.}
\label{tab:throughput}
\vspace{-0.05in}
\end{table}

%% file: tables/runtime_analysis.tex



\begin{table}[!tb]
\begin{center}
\begin{small}
\begin{tabular}{l|lcc}
\toprule
\multicolumn{2}{l}{LookupFFN}   & naive & opt1 \\ 
\midrule
\parbox[t]{2mm}{\multirow{6}{*}{\rotatebox[origin=c]{90}{Gather}}} & Latency          & 100 ms            & 35 ms            \\ 
& Compute Utilization & 1.21\%          & 3.53\%             \\
& Sustained L1 BW     & 79.7 GB/s        & 231.5 GB/s           \\ 
& Sustained LLC BW    & 9.5 GB/s         & 52.5 GB/s           \\ 
& L1 Miss \%          & 11.87\%           & 22.68\%             \\
& LLC Miss \%         & 69.56\%           & 12.77\%             \\ \midrule
\multicolumn{2}{l}{Hash Latency}  & 208 ms        & {106 ms}      \\ 
\multicolumn{2}{l}{Other Latency}  & 20 ms        & 20 ms      \\ \midrule
\multicolumn{2}{l}{Total Latency} & 328 ms        & {160 ms}            \\ \bottomrule
\end{tabular}
\end{small}
\end{center}
\vspace{-0.1in}
\caption{Analysis of Performance Characteristics for Lookup FFN. We keep volume of work and working set constant for both so instruction count and FLOPs are constant.}
\label{tab:analysis}
\vspace{-0.1in}
\end{table}

%% file: figures/speedup_opp.tex
\begin{figure}[!tb]
\centering
\includegraphics[width=2.5in]{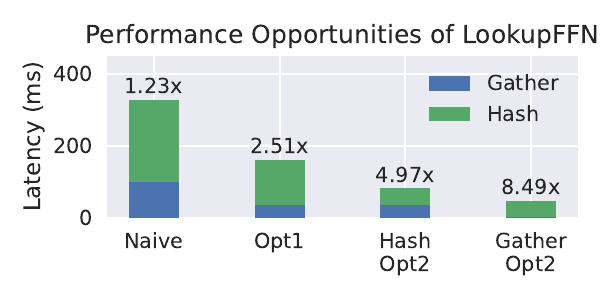}
\vspace{-0.16in}
\caption{Future performance opportunities of Lookup FFN. Each bar shows the time breakdown of LookupFFN by GatherAdd and Hash performance. Each version of LookupFFN is annotated with its speedup over Vanilla FFN.}
\label{fig:speedup_opp}
\vspace{-0.2in}
\end{figure}

%% file: sections/conclusions.tex
We conclude with a contemplative remark followed by a practical comment. 
Given the rapid improvements in compute capabilities of GPUs we have seen over the last decade (and a mature software support for different kernel shapes), there was little incentive for algorithm designers to use non-GEMM operations. In fact, any modern deep 
learning algorithm that is not heavily reliant 
on GEMM may 
have little chance of broader adoption, assuming it even sees the light of day. The ideas described here (and in Slide, Mongoose, YOSO), in some sense, lie at the other extreme. LookupFFN almost operates as if GEMM is forbidden. While compute capability 
improvements are slowing down, new memory technologies 
are already available and others in the development pipeline, we 
hypothesize that novel yet-undiscovered 
DNN architectures must hit a sweet-spot to delicately 
balance the trade-off between these two resources. 
Doing so also offers other benefits including potential energy savings. Now, specific to the model in this work, we  expect that the LookupFFN benefits can translate 
to other DNN models. This will complement the server chip developments in \S\ref{sec:introduction}, and enable AI models to serve a broader cross-section of industries.

%% file: sections/acknowledgement.tex
This work was supported in part by funding from the Vilas Board of Trustees and UW–Madison Office of the Vice Chancellor for Research and Graduate Education.